\pdfoutput=1

\documentclass[11pt]{article}

\usepackage{acl}

\usepackage{times}
\usepackage{latexsym}

\usepackage[T1]{fontenc}

\usepackage[utf8]{inputenc}

\usepackage{microtype}

\usepackage{inconsolata}
\usepackage{graphicx}
\usepackage{booktabs}

\title{Self-Prompt Tuning: Enable Autonomous Role-Playing in LLMs}

\author{\quad Aobo Kong$^{1}$ \quad Shiwan Zhao$^{2}$ \quad Hao Chen$^{3}$ \quad Qicheng Li$^{1}$\thanks{~~Qicheng Li is the corresponding author.} \quad Yong Qin$^{1}$\\
\textbf{\quad Ruiqi Sun$^{3}$ \quad Xin Zhou$^{3}$ \quad Jiaming Zhou$^{1}$ \quad Haoqin Sun$^{1}$}\\
$^1$TMCC, CS, Nankai University \quad $^2$Independent Researcher\\
$^3$Enterprise \& Cloud Research Lab, Lenovo Research \\
\texttt{$^{1}$kongaobo@mail.nankai.edu.cn \quad $^{2}$zhaosw@gmail.com} \\
\texttt{$^{1}$\{liqicheng, qinyong\}@nankai.edu.cn}\\
\texttt{$^{3}$\{chenhao31, sunrq2, zhouxin16\}@lenovo.com}\\
}

\begin{document}
\maketitle
\begin{abstract}

Recent advancements in LLMs have showcased their remarkable role-playing capabilities, able to accurately simulate the dialogue styles and cognitive processes of various roles based on different instructions and contexts. Studies indicate that assigning LLMs the roles of experts, a strategy known as role-play prompting, can enhance their performance in the corresponding domains. However, the prompt needs to be manually designed for the given problem, requiring certain expertise and iterative modifications. To this end, we propose self-prompt tuning, making LLMs themselves generate role-play prompts through fine-tuning. Leveraging the LIMA dataset as our foundational corpus, we employ GPT-4 to annotate role-play prompts for each data points, resulting in the creation of the LIMA-Role dataset. We then fine-tune LLMs like Llama-2-7B and Mistral-7B on LIMA-Role. Consequently, the self-prompt tuned LLMs can automatically generate expert role prompts for any given question. We extensively evaluate self-prompt tuned LLMs on widely used NLP benchmarks and open-ended question test. Our empirical results illustrate that self-prompt tuned LLMs outperform standard instruction tuned baselines across most datasets. This highlights the great potential of utilizing fine-tuning to enable LLMs to self-prompt, thereby automating complex prompting strategies. We release the dataset, models, and code at this \href{https://anonymous.4open.science/r/Self-Prompt-Tuning-739E/}{url}.

\end{abstract}

\section{Introduction}

\begin{figure*}[t]
\centering
\includegraphics[width=1\textwidth,trim=1.6cm 4.5cm 2.4cm 3.5cm, clip]{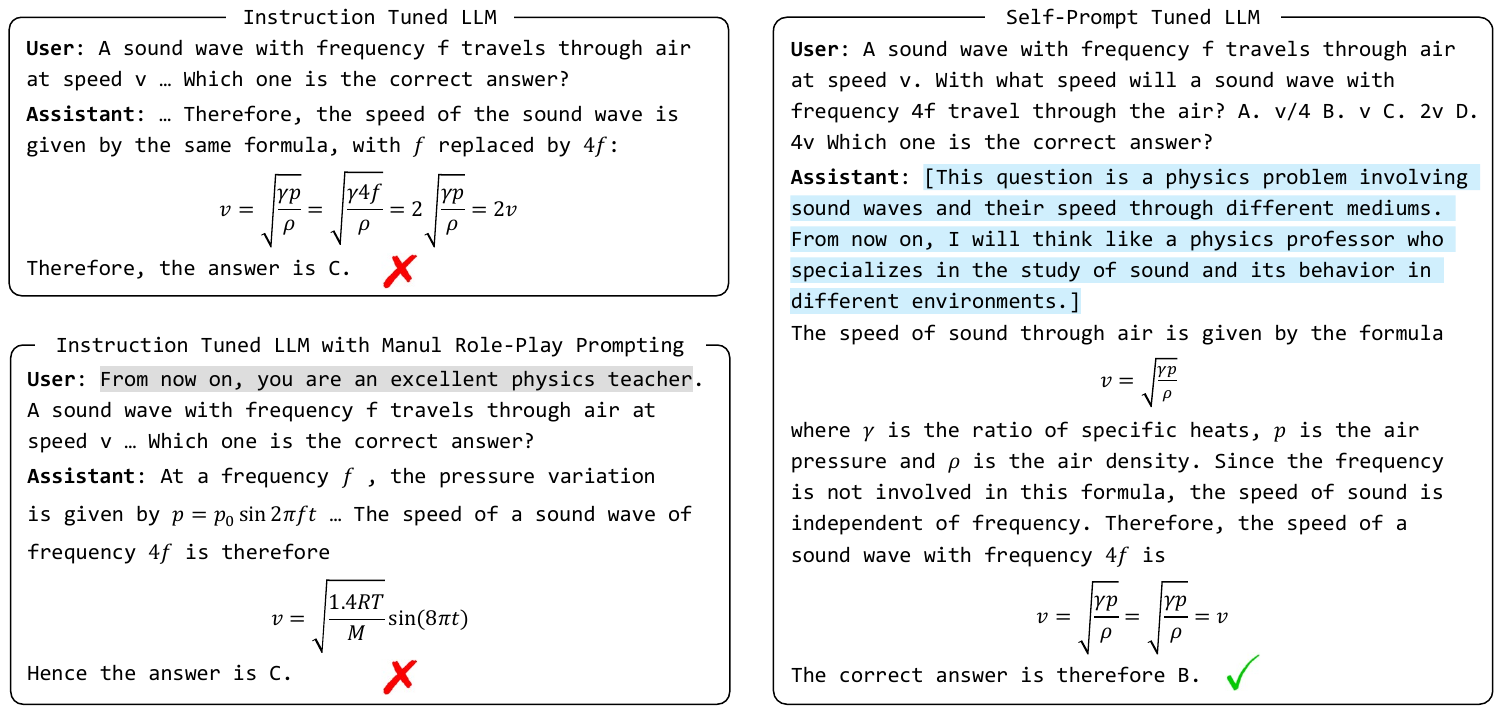} 
\caption{Examples of standard instruction tuned LLM, instruction tuned LLM with manual role-play prompting, and self-prompt tuned LLM on the same physics question. Manual and automatic role-play prompts are highlighted in gray and blue respectively. LLM used here is Mistral-7B.}
\label{fg: example}
\end{figure*}

Recent advances in large language models (LLMs) such as GPT-3 \cite{NEURIPS2020_1457c0d6}, PaLM \cite{chowdhery2022palm}, Llama \cite{touvron2023llama}, and Mistral \cite{jiang2023mistral} have dramatically reshaped the field of natural language processing (NLP). These models exhibit exceptional text understanding and generation capabilities, with performance that critically depends on the quality of the prompts used. To sufficiently unleash the potential of LLMs, a range of innovative prompting strategies have emerged. These include, but are not limited to, chain-of-thought prompting \cite{cot}, tree-of-thought prompting \cite{yao2023tree}, step-back prompting \cite{zheng2024step}, and the increasingly popular role-play prompting \cite{summarization, salewski2023incontext, kong2023better}. This paper concentrates on the development of self-prompt tuning to facilitate autonomous role-play prompting, a flexible method that may also be adapted for other prompting strategies.


Modern LLMs can seamlessly embody human characters\footnote{\href{https://beta.character.ai/}{Character.AI} offers LLMs impersonating celebrities, such as Albert Einstein.} and non-human entities\footnote{DeepMind researcher requires ChatGPT to act as a Linux terminal in the \href{https://www.engraved.blog/building-a-virtual-machine-inside/}{blog}.}, exhibiting incredible role-playing capabilities. While role-playing brings novel modes of interaction, it can also serve as a prompting strategy, termed role-play prompting, to enhance the performance of LLMs in various downstream NLP tasks. For instance, \citet{summarization} have LLMs impersonate judges with distinct personas and backgrounds to improve their summary assessment quality. In multi-domain QA tasks, \citet{salewski2023incontext} instruct LLMs to act as domain experts, leading to improved performance. Furthermore, \citet{kong2023better} assign diverse expert roles to LLMs more immersively through multi-turn dialogue, boosting their reasoning abilities. 
Despite its efficacy, role-play prompting faces two significant limitations common to many popular prompting strategies:

(\romannumeral1) It is task-specific. The role selection and prompt design must be tailored to individual tasks, and prompts are often not transferable to different tasks.

(\romannumeral2) The prompt design is labor-intensive, requiring significant domain expertise and iterative refinement, which can be time-consuming.


To address these limitations, could we leverage LLMs themselves to generate prompts, thereby reducing the reliance on human intervention? A natural idea is to utilize prompts to instruct models to generate prompts themselves. The NLP community has attempted to automatically situate LLMs in the appropriate role for the user across multiple rounds of dialogue guided by well-designed prompts\footnote{https://github.com/JushBJJ/Mr.-Ranedeer-AI-Tutor}. 
However, this prompt-based automation method tends to complicate the interaction process and introduce an excessive number of additional tokens, leading to diminished practicality.

While prompting strategies have positively modulate the behavior of LLMs in a cost-efficient manner, the pursuit of directly adjusting model parameters has led to the emergence of new methods like instruction tuning (IT) \cite{wei2022finetuned, wang-etal-2023-self-instruct, lima}. Through fine-tuning LLMs on a collection of datasets described via instructions, IT enables LLMs to follow huamn instructions without any additional prompts. 
Building on this foundation, this paper introduces \textbf{self-prompt tuning}, an innovative approach that enables LLMs to autonomously establish an appropriate role (i.e., role-play prompting) and respond accordingly through fine-tuning. Specifically, we leverage GPT-4 with in-context learning to reconstruct LIMA \cite{lima}, a small scale IT datasets, by adding corresponding role descriptions to each question. The resulting dataset is termed LIMA-Role. Subsequently, we fine-tune LLMs, such as Mistral-7B and Llama-2-7B, on this augmented dataset. The self-prompt tuned LLMs can automatically generate corresponding role-play prompts for a given question as shown in Figure \ref{fg: example}. 
We compare self-prompt tuned LLMs with instruction tuned baselines using 8 traditional benchmarks and an open-ended question test. Our results demonstrate consistent improvements over standard instruction tuned baselines on the majority of datasets, proving the efficacy of self-prompt tuning.

To the best of our knowledge, self-prompt tuning is the first to make LLMs themselves to generate prompts by fine-tuning. Our method opens a new avenue for automating diverse prompting strategies. We believe our work will catalyze further exploration in automating more advanced prompting techniques, 
such as least-to-most prompting \cite{zhou2023leasttomost} and tree-of-thought prompting \cite{yao2023tree}.


Our main contributions are as follows:
\begin{itemize}
\item  We propose self-prompt tuning, a novel approach achieving automation of role-play prompting through fine-tuning LLMs.
\item We release LIMA-Role, an enhanced version of the LIMA dataset annotated with role-play prompts using GPT-4, alongside LLMs fine-tuned on this dataset.
\item  We thoroughly evaluate self-prompt tuned LLMs using 8 traditional benchmarks and an open-ended question test, demonstrating the efficacy of self-prompt tuning.
\end{itemize}

\section{Related Work}

\subsection{Instruction Tuning}

Original pre-trained large language models (LLMs) excel as few-shot learners but struggle in zero-shot scenarios. \citet{wei2022finetuned} propose instruction tuning, a technique that fine-tunes LLMs on a diverse set of NLP datasets described via instructions, significantly improving their zero-shot performance. Following this approach, subsequent works like T0 \cite{sanh2022multitask}, FLAN-T5 \cite{chung2024scaling}, and ZeroPrompt \cite{xu-etal-2022-zeroprompt} expand the variety of tasks and the scale of data used for instruction tuning, further enhancing the models' capabilities. However, the data utilized in these works originated from traditional NLP datasets, which still lack diversity and complexity compared with real queries of human users. To solve this problem, researchers have attempted to leverage human annotators or LLMs to construct new datasets that better align with real-world human instructions. OpenAssistant \cite{pf2023openassistant} is an open-source assistant-style conversation corpus annotated by worldwide crowd-sourcing. Self-Instruct \cite{wang-etal-2023-self-instruct} generates 52k instruction-response pairs based on 175 manually-written prompts using LLMs. Evol-Instruct \cite{xu2024wizardlm} also relies on an initial set of instructions and employs LLMs to iteratively rewrite them into more complex instructions. LIMA \cite{lima} trains a LLM that approaches the capabilities of proprietary models using small-scale but high-quality data collected from wikiHow, Stack Exchange, and Reddit. Orca \cite{mukherjee2023orca} progressively fine-tunes LLMs on a massive corpus generated by GPT-4 to enhance their reasoning abilities. Essentially, instruction tuning alleviates the burden on users to craft prompts. And our proposed self-prompt tuning takes a further step by automating more complex prompting strategy.

\begin{figure*}[t]
\centering
\includegraphics[width=1\textwidth,trim=3.5cm 2.3cm 2.5cm 2.5cm, clip]{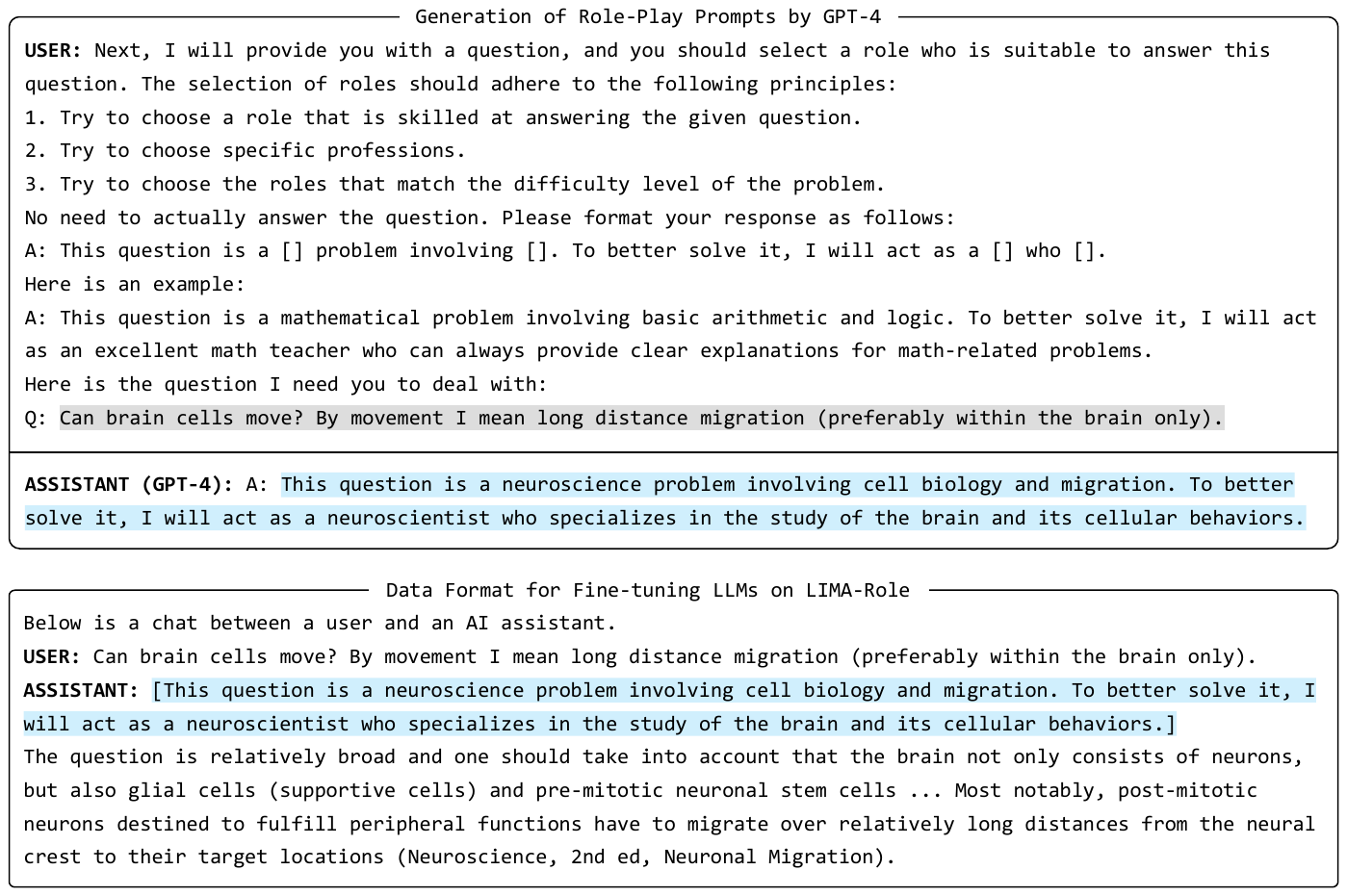} 
\caption{An illustration of LIMA-Role dataset construction process. The upper sub-image displays the prompt used for GPT-4 role-play prompt annotation. The lower sub-image shows how role-play prompts are utilized to construct LIMA-Role. The question to be annotated and the corresponding role-play prompts generated by GPT-4 are highlighted in gray and blue, respectively.}
\label{fg: framework}
\end{figure*}

\subsection{Role-playing Abilities of LLMs}

Modern LLMs exhibit remarkable adaptability and interactive capabilities in role-playing tasks. These models can flexibly adjust their output style according to the needs of different roles, providing users with a customized conversation experience. \citet{shanahan2023role} advocates LLMs as role simulators and warns against falling into the trap of anthropomorphism. \citet{wang2023rolellm} propose RoleLLM, a role-playing framework of data construction and evaluation. Beyond facilitating immersive interactions, role-playing can also enhance the model's performance across downstream NLP tasks. \citet{summarization} employ LLMs to emulate judges possessing unique personas and backgrounds, thereby enhancing the quality of their summarization assessments. \citet{salewski2023incontext} direct Large Language Models (LLMs) to embody domain-specific expertise, leading to enhanced performance in multi-domain QA tasks. \citet{kong2023better} immerse LLMs in diverse expert roles via multi-turn dialogues, thereby augmenting their reasoning capabilities. Role-play is also employed in LLM-based multi-agent frameworks \cite{park2023generative, xiong-etal-2023-examining, liang2023encouraging}. These studies utilize role-play prompting to facilitate the cooperative interaction among multiple agents. While the efficacy of role-play prompting has been demonstrated, the need of manually crafting prompts for each task hinders its broader application. To alleviate this bottleneck, we propose self-prompt tuning, a novel approach that automates prompt design by LLMs themselves, thereby minimizing human intervention.

\subsection{Prompting Strategies}

Extensive research and practice have demonstrated that prompts significantly impact the performance of LLMs. To fully unlock the potential of LLMs, various complex prompting strategies, not just role-play prompting, have been developed. Least-to-most prompting \cite{zhou2023leasttomost} decomposes the original problem into simpler subproblems and solves them in sequence. Self-refine prompting \cite{madaan2023selfrefine} generates an output first and then employs the same LLM to provide feedback and refinement, iteratively improving the initial output. Tree-of-thought \cite{yao2023tree} prompting represents potential reasoning paths as a branching tree structure and utilizes search algorithms like DFS or BFS to explore and identify the correct reasoning path. Step-back prompting \cite{zheng2024step} involves abstracting information to derive high-level concepts and first principles, which are then utilized to guide the reasoning process. These prompting strategies necessitate providing few-shot examples to guide LLMs in following a specific thought pattern. Our proposed self-prompt tuning introduces a novel approach that involves constructing a dataset embodying the desired thought process and then fine-tuning LLMs to inject this thinking pattern into their parameters. Our experiments have demonstrated the success of this method in role-play prompting. And we leave the extension of self-prompt tuning in other prompting strategies to future work.

\section{Self-Prompt Tuning}

In this section, we introduce our proposed self-prompt tuning in detail. Self-prompt tuning consists of two steps as follows: (1) Modify an existing instruction tuning dataset to include role-play prompts. (2) Fine-tune LLMs on the resulting dataset to enable them automatically generate role-play prompts tailored to the specific questions.

\subsection{Construct LIMA-Role Dataset}

The small scale yet high-quality instruction tuning dataset, LIMA \cite{lima}, comprises 1,000 single-turn dialogues and 30 multi-turn dialogues,  making it highly suitable to serve as a foundational dataset. Studies by \citet{salewski2023incontext, kong2023better} demonstrate that taking on expert roles for a given task can typically enhance the model's performance. Building on this premise, we employ GPT-4 in one-shot manner to generate expert role-play prompts for each training instance in LIMA (only consider the first question for multi-turns data). These role-play prompts are then prefixed to the corresponding answers, yielding a new dataset, LIMA-Role. Inspired by chain-of-thought prompting \cite{cot}, the question summarization is also designed into the role-play prompt, aiming to help generate correct role descriptions. We provide prompts utilized for GPT-4 and an example illustrating the process of modifying one data instance in Figure \ref{fg: framework}. Additionally, GPT-4 declines to generate role prompts to some unsafe, biased or unethical questions in LIMA, 14 in total. We manually design prompts with the role of "AI assistant" for these questions. 

While LLMs have demonstrated remarkable capabilities in data annotation tasks \cite{wang-etal-2023-self-instruct, xu2024wizardlm, xu-etal-2023-baize}, it remains necessary to validate the data quality of LIMA-Role. We conduct a random selection of 100 entries from the dataset to undergo manual evaluation, focusing on three key aspects: formatting, question summarization, and role description. The assessment reveals that 100\% of the entries maintain a consistent format, 96\% correctly summarize the questions, and 97\% offer appropriate role descriptions. Therefore, we conclude that the data quality of LIMA-Role meets our criteria.

\subsection{Fine-tune LLMs on LIMA-Role}

After completing the construction of LIMA-Role, we fine-tune original pre-trained LLMs like Mistral-7B on that dataset with the standard supervised loss. We organize the data in the form of interaction between "AI assistant" and "user", and set a fixed system prompt, as shown in Figure \ref{fg: framework}.

\section{Experiments}

\begin{table*}[!t]
\centering
\scalebox{0.89}{
\begin{tabular}{lccccccccc} 
\toprule
Model                 & MMLU          & CSQA & Strategy & Truthful    & OpenBook      & HumanEval     & GSM8K         & Date          & AVG   \\ 
\midrule
ChatGPT               & 67.3          & 76.9    & 61.7     & 60.2          & 81.6             & 68.3             & 80.8          & 67.8             & 70.6     \\ 
\midrule
\multicolumn{10}{c}{Llama-2-7B}                                                                                                                \\ 
\midrule
Llama-Chat            & 44.0             & 58.6    & 59.0        & 40.4             & 63.6             & 13.7             & 29.3             & 49.3             & 44.7     \\ 
\midrule
Llama-LIMA            & 40.4             & 48.6    & 55.5        & 39.7             & 48.2             &  9.4         & 13.5             & 43.1             & 37.3     \\
Llama-Role            & \textbf{42.9}             & \textbf{57.3}    & \textbf{59.5}        & \textbf{47.8}             & \textbf{52.1}             & 8.7             & \textbf{13.6}            & \textbf{43.1}             & \textbf{40.6}     \\ 
\midrule
Llama-LIMA$^\dagger$           & 41.8             & 49.5    & 57.2        & 38.9             & \textbf{50.6}             &  \textbf{9.4}         & 14.0             & \textbf{44.2}             & 38.2     \\
Llama-Role$^\dagger$             & \textbf{44.1}             & \textbf{58.0}    & \textbf{59.6}        & \textbf{48.0}             & 50.2             & 8.5             & \textbf{14.5}            & 42.8             & \textbf{40.7}     \\ 
\midrule
\multicolumn{10}{c}{Mistral-7B}                                                                                                                 \\ 
\midrule
Mistral-Instruct & 51.1          & 66.4    & 60.2     & 51.8          & 72.2             & 33.2             & 35.2          & 56.4             & 53.3     \\ 
\midrule
Mistral-LIMA          & 53.2          & 52.6 & 58.5     & 43.9          & 63.1          & 25.9          & 22.4          & 40.6          & 45.0  \\
Mistral-Role          & \textbf{56.0} & \textbf{59.8} & \textbf{61.9}     & \textbf{46.1} & \textbf{68.2} & \textbf{26.6} & \textbf{25.8} & \textbf{42.7} & \textbf{48.4}  \\
\midrule
Mistral-LIMA$^\dagger$           & 53.4          & 54.8 & 59.3     & 42.7          & 63.4          & \textbf{27.9}          & 20.4          & \textbf{42.5}          & 45.6  \\
Mistral-Role$^\dagger$           & \textbf{57.1} & \textbf{61.3} & \textbf{62.8}     & \textbf{45.3} & \textbf{69.6} & 27.8 & \textbf{27.1} & 42.0 & \textbf{49.1}  \\
\bottomrule
\end{tabular}}
\caption{The performance of self-prompt tuned LLMs, standard instruction tuned LLMs (LIMA version and official version), and ChatGPT on each dataset. Without $\dagger$: average performance of the four models. With $\dagger$: results from the model with the best average performance among the four models.}
\label{tb: results}
\end{table*}

\subsection{Tasks and Datasets}
Initial investigations into instruction tuning \cite{lima, xu2024wizardlm} involved comparing various LLMs' responses to open-ended questions, utilizing both human and GPT-4 assessments to gauge their quality. \citet{gudibande2024the} highlighted that relying solely on this evaluation method may result in an overestimation of model quality. Therefore, we combine traditional NLP benchmarks and open-ended questions to comprehensively evaluate the efficacy of self-prompt tuning.

\vspace{0.08cm}

\noindent{\bf NLP Benchmarks} We hope that self-prompt tuned LLMs can automatically generate expert role-play prompts for different questions. Therefore, datasets containing multi-domain problems are highly suitable for evaluation. MMLU \cite{hendrycks2021measuring} is a multi-domain QA dataset and has been widely used to evaluate LLMs. We sample 2000 questions from MMLU, balanced across 10 categories (35 subcategories). CSQA \cite{talmor-etal-2019-commonsenseqa}, StrategyQA \cite{10.1162/tacl_a_00370}, TruthfulQA \cite{lin-etal-2022-truthfulqa}, and OpenBookQA \cite{mihaylov-etal-2018-suit} are also muti-domain datasets and included. We additionally add GSM8K (math) \cite{cobbe2021training}, HumanEval (code) \cite{chen2021evaluating}, Date Understanding (reasoning) \cite{srivastava2023beyond} to enrich the form and content of the evaluation. More details can be found in Table \ref{tb: dataset}.

\begin{table}[h]
\centering
\begin{tabular}{lccc} 
\toprule
Dataset            & $N_{q}$   & $L_{q}$   & Format         \\ 
\midrule
MMLU               & 2000 & 79.4 & option (A-D)   \\
CSQA               & 1221 & 27.8 & option (A-E)   \\
StrategyQA         & 2290 & 9.6  & yes or no      \\
TruthfulQA         & 817  & 47.3 & option (A-D)   \\
OpenbookQA         & 500  & 26.5 & option (A-D)   \\
HunamEval          & 164  & 67.7 & code           \\
GSM8K              & 1319 & 46.9 & arabic number  \\
Date  & 369  & 35.0 & Option (A-F)   \\
\midrule
LIMA-Test & 300 & 21.3 & free \\
\bottomrule
\end{tabular}
\caption{Relevant information of benchmarks and LIMA test set. $N_{q}$ denotes the number of questionsin each dataset. $L_{q}$ denotes the average words of questions in each dataset. Format denotes the answer format of each dataset.}
\label{tb: dataset}
\end{table}

\vspace{0.08cm}

\noindent{\bf Open-ended Questions}
We leverage the LIMA test set, comprising 300 challenging questions authored by real users, to assess the capabilities of LLMs. See more details in Table \ref{tb: dataset}.

\subsection{Experimental Setup}

\noindent{\bf Models}
We self-prompt tune original Mistral-7B and Llama-2-7B, which are the leading open-source LLMs at the time of writing.

\vspace{0.05cm}

\noindent{\bf Baselines}
In addition to comparing self-prompt tuned LLMs on LIMA-Role and instruction tuned LLMs on original LIMA, we also present the experimental results of ChatGPT (gpt-3.5-turbo-0125), Llama-2-chat (the official version), and Mistral-instruct (the official version) to enhance our comprehension of the models' capabilities.

\vspace{0.08cm}

\noindent{\bf Training Details}
In line with prior research \cite{lima}, we respectively conduct fine-tuning of Mistral-7B on LIMA and LIMA-Role datasets for 4 epochs, employing AdamW optimization with parameters $\beta_{1}=0.9$, $\beta_{2}=0.999$, and a weight decay of 0.1. We initialize the learning rate to $1e-5$ without warmup, implementing a cosine decay schedule that decreases to 0 by the end of training. The batch size is set to 64, with a maximum token limit of 4096. To mitigate overfitting, dropout is applied to attention calculations, starting at $p_{d}$ = 0.0 at the bottom layer and linearly raising the rate to $p_{d}$ = 0.25 at the last layer. We utilize FlashAttention-2 \cite{dao2024flashattention} to optimize memory usage and expedite training. The method and parameter settings for fine-tuning Llama-2-7B mirror those of Mistral-7B, differing only in the number of training epochs, which is set to 8. Training is performed on 4 A100-80G. Due to the small data scale of LIMA dataset, model performance exhibits variability; hence, we fine-tune four models for the same dataset using different seeds and average their performance across NLP benchmarks.

\vspace{0.06cm}
\noindent{\bf Evaluation Details}
For both NLP benchmarks and the LIMA test set, evaluations are conducted in a zero-shot manner, without any few-shot exemplars. Consistent with prior studies \cite{NEURIPS2022_8bb0d291, kong2023better}, we employ greedy decoding with a temperature of 0 to ensure deterministic results. While averaging the performance of four models fine-tuned on the same dataset across NLP benchmarks, we select the model with the best average performance from the four and evaluate it on the LIMA test set. The quality of their responses is assessed using GPT-4 (gpt-4-1106-preview, we adopt the prompt proposed by \citet{lima}). Role-play prompts generated by self-prompt tuned LLMs are invisible to GPT-4 to ensure fairness.

\subsection{Results on NLP Benchmarks}

\begin{figure*}[t]
\centering
\scalebox{0.87}{
\includegraphics[width=1\textwidth,trim=2cm 5cm 2cm 4.5cm, clip]{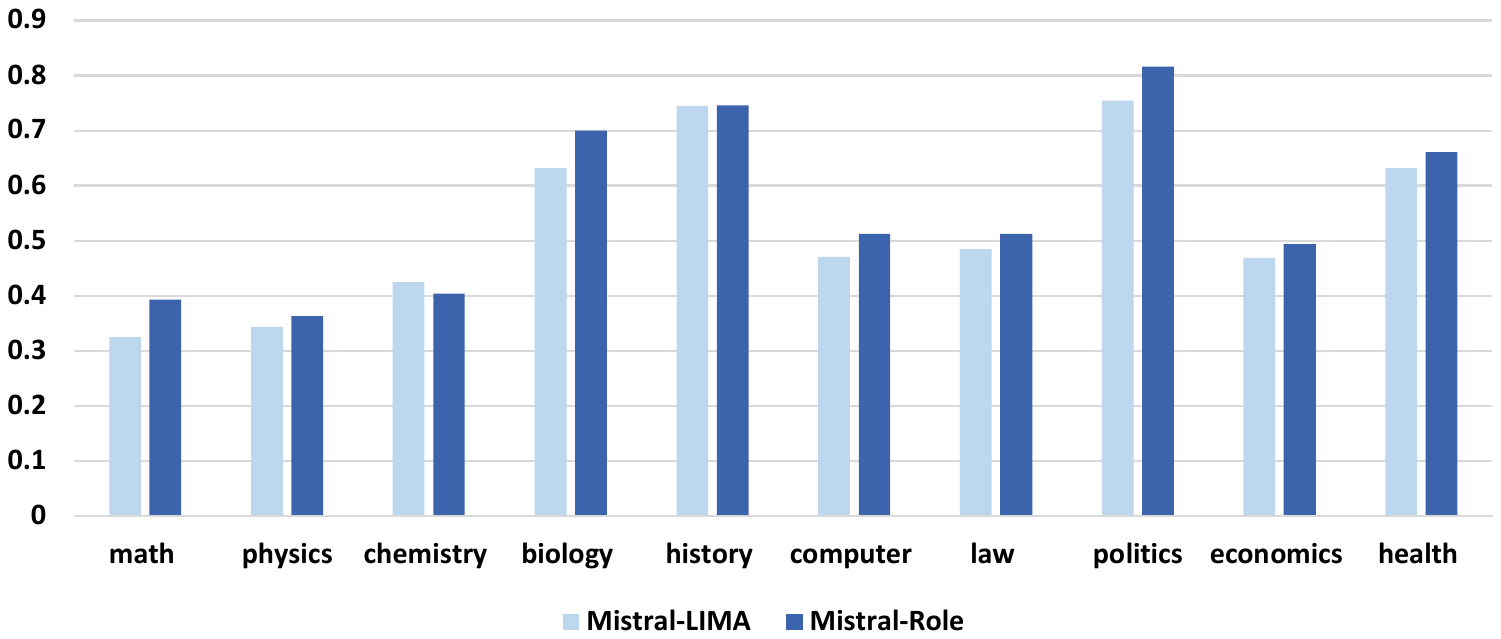} }
\caption{The performance comparison between Mistral-LIMA and Mistral-Role across various domain-specific subsets in MMLU. Mistral-Role outperforms Mistral-LIMA in 9 out of 10 domains and underperforms in chemistry.}
\label{fg: mmlu}
\end{figure*}

\begin{figure*}[t]
\centering
\scalebox{1}{
\includegraphics[width=1\textwidth,trim=0cm 5.5cm 0cm 4.5cm, clip]{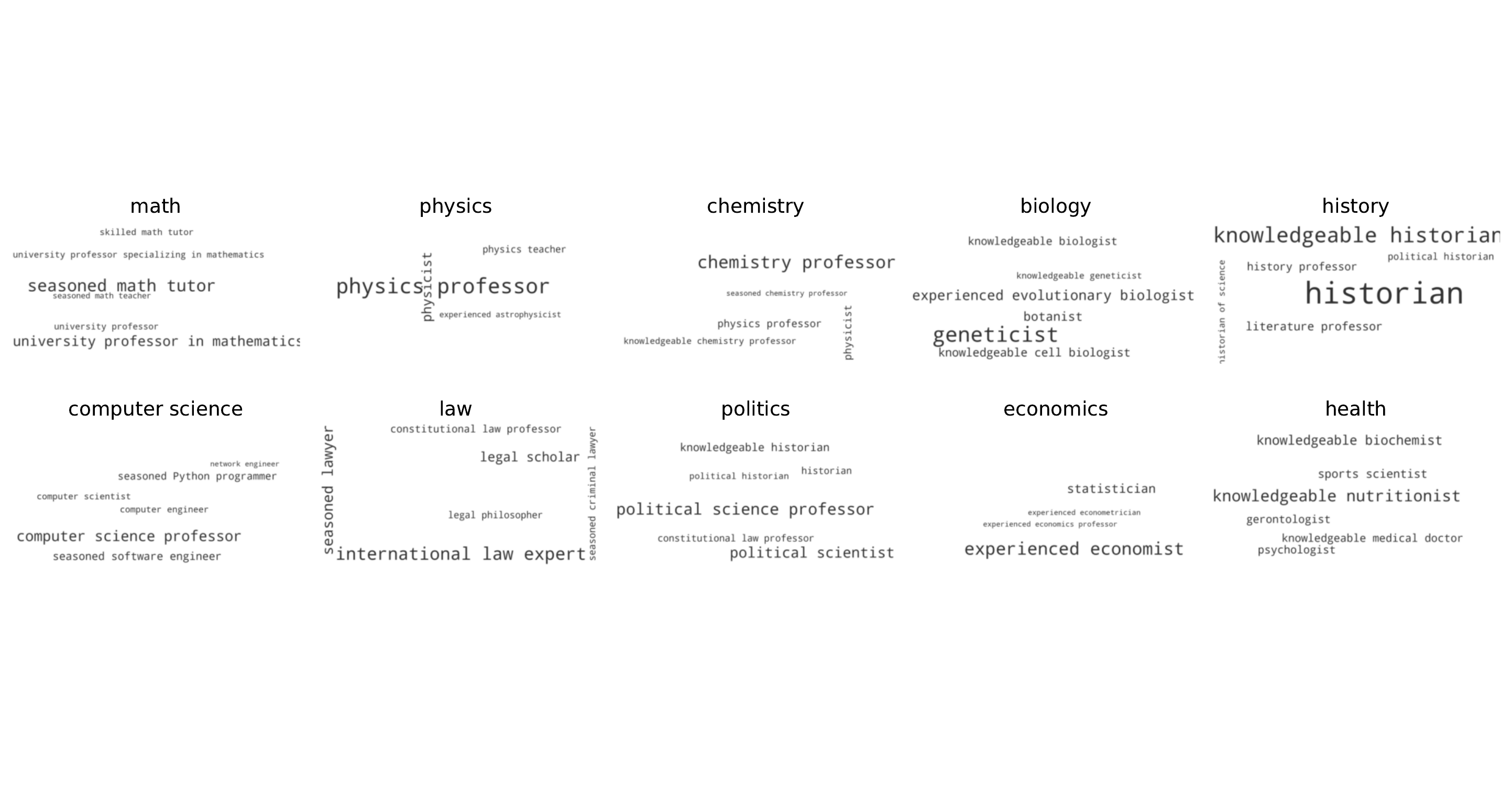} }
\caption{Word clouds based on roles generated by Mistral-Role across domain-specific subsets in MMLU. Words characterized by larger font sizes and deeper color correspond to higher frequencies.}
\label{fg: cloud}
\end{figure*}

Detailed experimental results on NLP benchmarks are presented in Table \ref{tb: results}. We report both the average performance and peak performance of LLMs simultaneously. For HumanEval, the evaluation metric utilized is pass@1, whereas accuracy serves as the metric for the remaining datasets.

\noindent{\bf Average Performance Comparison}
As shown in Table \ref{tb: results}, self-prompt tuned LLMs consistently outperform those instruction-tuned on LIMA across the majority of benchmarks, demonstrating the efficacy of our approach. Delving deeper, we compare the performance of Mistral-Role and Mistral-LIMA on domain-specific subsets within the MMLU. According to the results in Figure \ref{fg: mmlu}, Mistral-Role outperforms Mistral-LIMA in 9 out of 10 domains (28 out of 34 subcategories) revealing that self-prompt tuning is beneficial across a diverse range of fields. Moreover, to assess the capability of self-prompt tuned LLMs to automate role-play prompting, we extract roles automatically generated by Mistral-Role for questions in each  domain-specific subset in MMLU. By identifying and visualizing the most frequent roles through word clouds in Figure \ref{fg: cloud}, we observe that Mistral-Role assigns appropriate expert roles to questions across different domains. This highlights that self-prompt tuning successfully enables LLMs to autonomously generate role-play prompts. We also observe that self-prompt tuned LLMs exhibit unstable performance improvement on single-domain tasks compared to multi-domain QA tasks (Llama-Role on HumanEval, GSM8K, and Date). \citet{kong2023better} reveal that while expert roles generally brings performance gains, this improvement is not guaranteed. In single-domain tasks, where the format of questions tends to be highly consistent, the role-play prompts generated by self-prompt tuned LLMs are quite similar. This lack of diversity in the prompts likely contributes to the observed instability in performance improvements. Conversely, for multi-domain QA tasks, the diversity in the generated role-play prompts is notably higher, leading to stable improvement. Thus, the limited improvement of Llama-Role in single-domain tasks can be attributed to this factor.

\begin{table*}
\centering
\scalebox{0.9}{
\begin{tabular}{clcc} 
\toprule
No. & Prompt                                                                             & MDQA   & SDTask  \\ 
\midrule
0   & None                                                                               & 54.3 & 29.6   \\
\midrule
1   & {[}Question Description].                                                          & 53.8 & 29.5   \\
2   & {[}Question Description]. As a result, I will solve it like [Role Description].    & 57.3 & 31.0   \\
3   & {[}Question Description]. Therefore, I will answer it as [Role Description].       & 57.4 & 31.8   \\
4   & {[}Question Description]. To solve this problem, I will act as [Role Description]. & 57.9 & 24.7      \\
5   & {[}Question Description]. So I will become [Role Description].                     & 58.6 & 31.3   \\
6   & {[}Question Description]. Fortunately, I am [Role Description].                    & 58.4 & 32.9   \\
7   & {[}Question Description]. For this reason, I will be [Role Description].           & 57.4 & 30.6   \\
8   & {[}Question Description]. From now on, I will think like [Role Description].       & 58.4 & 31.7   \\
\bottomrule
\end{tabular}}
\caption{The performance of Mistral-Role adopting different prompt designs. Similarly, we train four models for each prompt design with different random seeds and report the average performance here.}
\label{tb: ablation}
\end{table*}

\vspace{0.08cm}
\noindent{\bf Peak Performance Comparison}
Self-prompt tuned LLMs with the best average performance still surpass standard instruction tuned baselines as indicated in Table \ref{tb: results}. However, when comparing with official instruction-tuned versions, the self-prompt tuned LLMs tend to underperform. It's crucial to emphasize that both Llama-Role and Mistral-Role are fine-tuned on only 1030 data points, whereas the official versions are fine-tuned on datasets exceeding 10,000 data points and undergo complex RLHF \cite{rlhf}. This discrepancy in training dataset scale and methodology accounts for the performance differences observed.

\subsection{Results on Open-ended Questions}
\begin{figure}[t]
\centering
\scalebox{0.47}{
\includegraphics[width=1\textwidth,trim=2.3cm 3.8cm 6cm 3.5cm, clip]{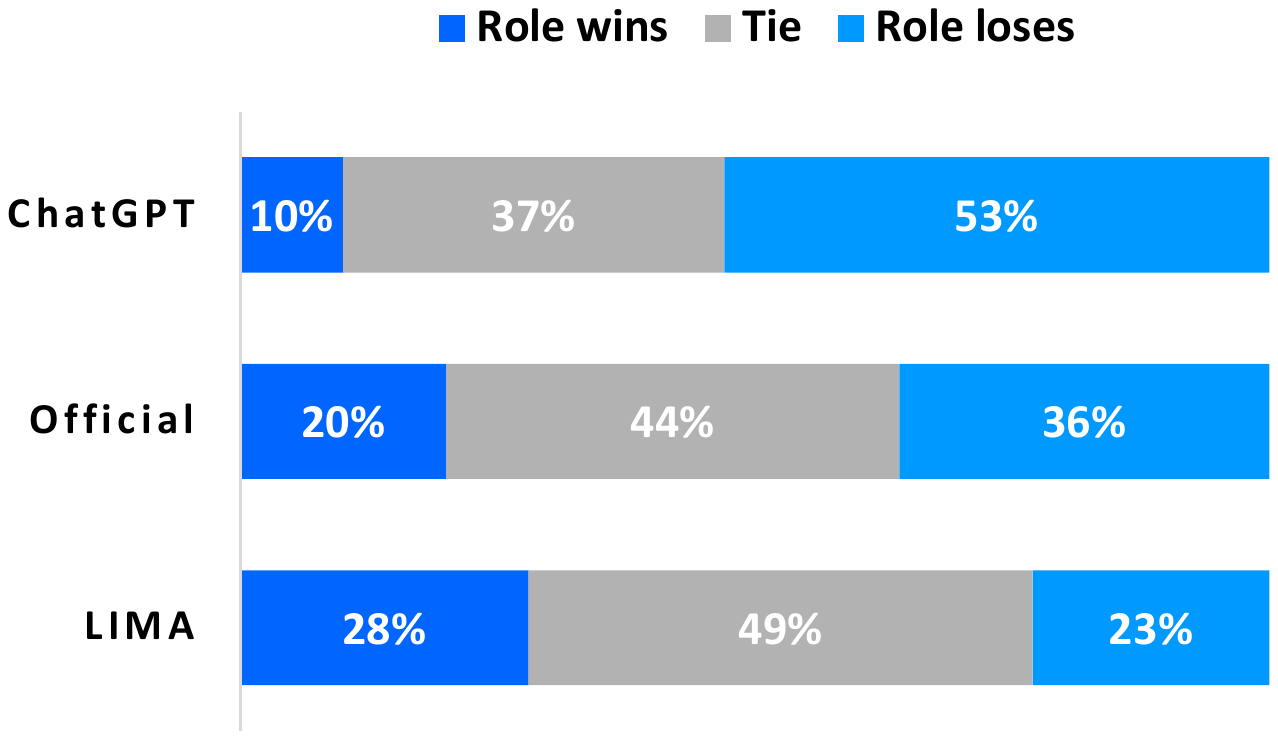} }
\caption{Preference evaluation on LIMA test set using GPT-4 as the annotator. In this context, LIMA refers to Mistral-LIMA, while Role denotes Mistral-Role.}
\label{fg: gpt-4}
\end{figure}

We select self-prompt tuned and standard instruction tuned Mistral-7B with the best average performance to conduct open-ended question test. Results annotated by GPT-4 are depicted in Figure \ref{fg: gpt-4}. Despite only inserting non-substantive role-play prompts into the LIMA dataset, Mistral-Role still generate better responses than Mistral-LIMA 5\% of the time, further underscoring the widespread effectiveness of self-prompt tuning. Nonetheless, Mistral-Role exhibits subpar performance compared to the official version and ChatGPT, indicating that merely 1,030 high-quality data points are insufficient for effectively fine-tuning a 7B-parameter model.

\subsection{Ablation Study}
While the performance of LLMs is highly sensitive to the prompt in various prompting strategies, the influence of prompt design on fine-tuning models remains unexplored. Given the high cost of accessing GPT-4, we maintain the question description and role description, only modifying the left sections of the prompt. The prompts we design and their practical results on Mistral are summarized in Table \ref{tb: ablation}. Prompt 1, containing only the question description, achieves the lowest performance, thereby eliminating interference from question descriptions. Prompts 2-8, which add role descriptions with variations at the junctions, consistently show improvements in both multi-domain QA tasks and single-domain tasks. Among these, Prompts 6 and 8 exhibit relatively optimal performance. We ultimately select Prompt 8, which demonstrates the most balanced performance improvement across each dataset, as the final design. The results indicate that prompt design also impacts the performance of fine-tuning LLMs, but not as sensitively as in non-fine-tuning scenarios.

\section{Conclusion}

In this paper, we propose self-prompt tuning, a novel approach that enables large language models (LLMs) to autonomously generate role-play prompts through fine-tuning. By first constructing the LIMA-Role dataset, which augments the LIMA dataset with expert role-play prompts generated by GPT-4, and then fine-tuning LLMs on this dataset, self-prompt tuned LLMs gained the ability to automatically generate relevant expert role-play prompts tailored to any given question. Comprehensive evaluations on 8 traditional NLP benchmarks and an open-ended question test reveal that self-prompt tuned LLMs consistently outperform standard instruction tuned baselines across the majority of datasets. The results highlight the efficacy of self-prompt tuning in automating role-play prompting. Overall, this work paves a promising new path for automating diverse complex prompting strategies.

\section*{Limitations}

Due to its small scale and ease of modification, we select the LIMA dataset as the foundational dataset. However, the data scale of 1,030 samples is insufficient to fully fine-tune a 7B parameter model, rendering our models unable to make a meaningful performance comparison with ChatGPT and the official versions. Moreover, we only manually make limited attempts at designing role-play prompts for the LIMA-Role dataset, and cannot guarantee that the optimal effects of self-prompt tuning were achieved. Last, owing to limited computational resources, we are unable to apply our method on LLMs with larger parameter scales. Consequently, we could not obtain conclusions about how the effects of self-prompt tuning vary as the scale of model parameters increases.



\bibliography{custom}




\end{document}